\documentclass[sigconf,screen]{acmart}
\renewcommand\footnotetextcopyrightpermission[1]{}

\usepackage{multirow}
\usepackage{multicol}
\usepackage{algorithm}
\usepackage[noend]{algpseudocode}
\usepackage{color}
\usepackage{colortbl}
\definecolor{Red}{cmyk}{0,1,1,0}
\definecolor{Green}{cmyk}{1,0,1,0}
\definecolor{Cyan}{cmyk}{1,0,0,0}
\definecolor{Purple}{cmyk}{0.45,0.86,0,0}
\definecolor{Rosolic}{cmyk}{0.00,1.00,0.50,0}
\definecolor{Blue}{cmyk}{1.00,1.00,0.00,0}
\definecolor{Orange}{cmyk}{0,0.52,0.80,0}
\definecolor{Black}{cmyk}{1,0,0,1}
\newcommand{\best}[1]{\cellcolor{red!30}#1}
\newcommand{\second}[1]{\cellcolor{orange!30}#1}
\newcommand{\third}[1]{\cellcolor{yellow!30}#1}

\AtBeginDocument{%
  }

\acmConference[ACM MM'25]{}{Dublin}{Ireland}

\begin{document}

\title{LL-Gaussian: Low-Light Scene Reconstruction and Enhancement via Gaussian Splatting for Novel View Synthesis}

\thanks{$^{\dagger}$Corresponding author.}
\author{Hao Sun$^{1,2}$, Fenggen Yu$^{4}$, Huiyao Xu$^{3}$, Tao Zhang$^{5}$, Changqing Zou$^{1,3\dagger}$} 
	 \affiliation{%
        \vspace{1em}
		   \institution{$^{1}$Zhejiang Lab \ \ \ \ $^{2}$University of Chinese Academy of Sciences \\ 
           $^{3}$State Key Lab of CAD\&CG, Zhejiang University \\ 
           $^{4}$Simon Fraser University \ \ \ \ $^{5}$Hangzhou Dianzi University}
           \country{}
           \city{}
		 }
\renewcommand{\shortauthors}{Sun et al.}

\begin{abstract}
Novel view synthesis (NVS) in low-light scenes remains a significant challenge due to degraded inputs characterized by severe noise, low dynamic range (LDR) and unreliable initialization. While recent NeRF-based approaches have shown promising results, most suffer from high computational costs, and some rely on carefully captured or pre-processed data—such as RAW sensor inputs or multi-exposure sequences—which severely limits their practicality. In contrast, 3D Gaussian Splatting (3DGS) enables real-time rendering with competitive visual fidelity; however, existing 3DGS-based methods struggle with low-light sRGB inputs, resulting in unstable Gaussian initialization and ineffective noise suppression. To address these challenges, we propose LL-Gaussian, a novel framework for 3D reconstruction and enhancement from low-light sRGB images, enabling pseudo normal-light novel view synthesis. Our method introduces three key innovations: 1) an end-to-end \textbf{L}ow-\textbf{L}ight \textbf{G}aussian \textbf{I}nitialization \textbf{M}odule (\textbf{LLGIM}) that leverages dense priors from learning-based MVS approach to generate high-quality initial point clouds; 2) a dual-branch Gaussian decomposition model that disentangles intrinsic scene properties (reflectance and illumination) from transient interference, enabling stable and interpretable optimization; 3) an unsupervised optimization strategy guided by both physical constrains and diffusion prior to jointly steer decomposition and enhancement. Additionally, we contribute a challenging dataset collected in extreme low-light environments and demonstrate the effectiveness of LL-Gaussian. Compared to state-of-the-art NeRF-based methods, LL-Gaussian achieves up to \textbf{2,000$\times$ faster} inference and reduces training time to just \textbf{2\%}, while delivering superior reconstruction and rendering quality.
\end{abstract}



\begin{teaserfigure}
  \setlength{\abovecaptionskip}{0.1cm}
  \includegraphics[width=\textwidth]{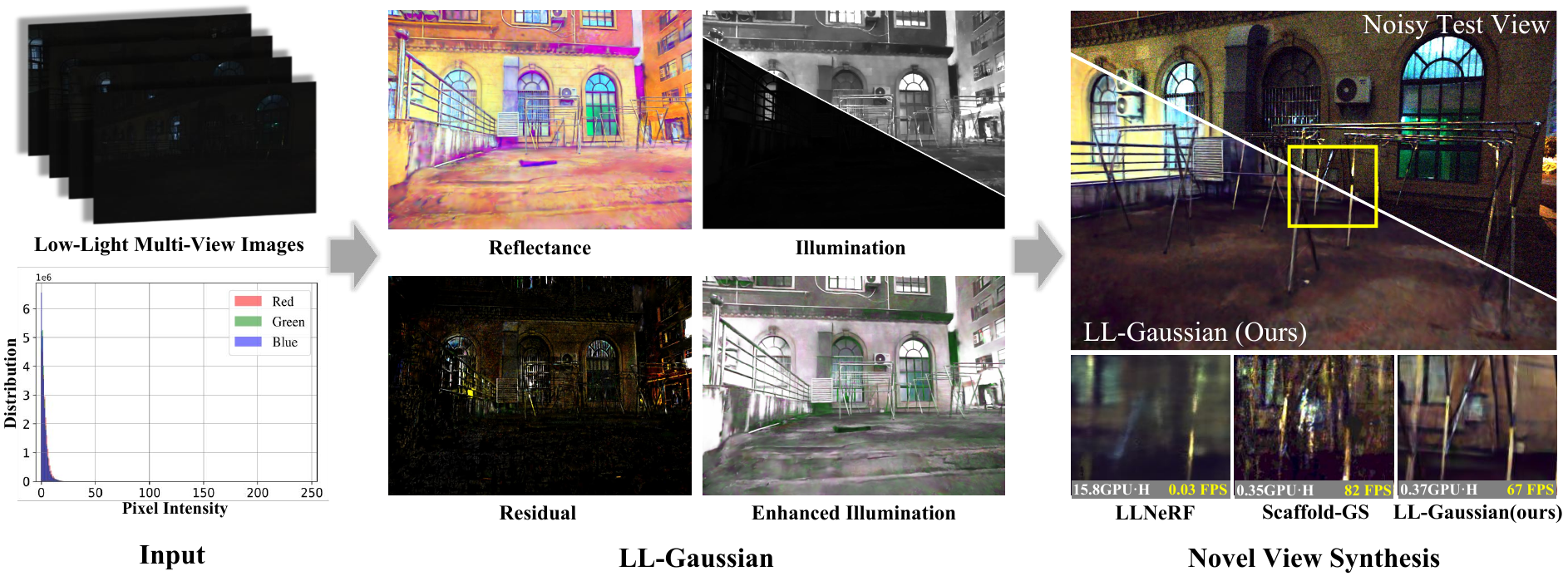}
  \caption{Given unposed multi-view images of extreme low-light scenes (Left), LL-Gaussian can decompose the scenes into reflectance, illumination and residual components, while learning an enhanced illumination field (Mid). Our method enables photorealistic normal-light novel view synthesis with strong robustness to sensor noise (Right). Compared to SOTA NeRF- and 3DGS-based baselines, LL-Gaussian achieves superior visual quality, efficient training process and real-time rendering. }
  \Description{}
  \label{fig:teaser}
\end{teaserfigure}

\maketitle

\section{Introduction}

Real-world 3D scene reconstruction and novel view synthesis (NVS) are fundamental tasks in computer vision with broad applications in autonomous driving, AR/VR, and robotics. Recent advances in neural scene representations, especially NeRF-based methods~\cite{barron2021mip,barron2022mip,muller2022instant,yu2021pixelnerf}, along with the emergence of 3D Gaussian Splatting (3DGS)~\cite{kerbl20233d,yu2024mip,lu2024scaffold,huang20242d,yu2024gaussian}, have significantly improved rendering quality and efficiency in well-lit environments. However, these methods predominantly assume high-quality, well-posed and well-exposed inputs—an assumption that often breaks down in challengeing scenes such as low-light environments. In practice, low-light conditions are common in real-world settings like nighttime driving, indoor robotics, surveillance, and disaster response, where lighting is limited or unpredictable. These environments introduce substantial sensor noise, color distortions, and reduced texture details, making it particularly difficult to recover accurate geometry and photometric consistency. As a result, overcoming these challenges is not only technically demanding but also essential for safety-critical and real-world applications.


Recent progress has extended NeRF-based techniques to low-light novel view synthesis. RawNeRF~\cite{mildenhall2022nerf} pioneers the use of RAW sensor data to reconstruct high dynamic range (HDR) scenes, demonstrating NeRF’s robustness to zero-mean sensor noise. Subsequent works such as LLNeRF~\cite{wang2023lighting} and AlethNeRF~\cite{cui2024aleth} further improve reconstruction performance of low-light low dynamic range (LDR) by unsupervised decomposition frameworks. While significant progress has been made in rendering quality, these methods suffer from prohibitive training times and slow inference speeds.

In contrast, 3DGS~\cite{kerbl20233d} has emerged as a compelling alternative, offering fast training process and real-time rendering while maintaining high fidelity. However, under low-light conditions, its explicit nature introduces new challenges. Unlike NeRFs, which employ multilayer perceptrons (MLPs) that inherently act as low-pass filters suppressing high-frequency noise, 3DGSs tend to overfit sensor noise by fitting thin and unstable Gaussian primitives~\cite{li2024chaos}. Several methods~\cite{li2024chaos,jin2024lighting,cai2024hdr,singh2024hdrsplat,wu2024hdrgs} extend 3DGS to low-light HDR reconstruction using RAW sensor inputs, achieving impressive quality and speed. Others, such as Gaussian-DK~\cite{ye2024gaussian} and Cinematic Gaussians~\cite{wang2024cinematic}, utilize multi-exposure LDR sequences and metadata (e.g., exposure time, ISO) to improve radiance estimation. However, these approaches heavily rely on specialized acquisition setups—such as RAW sensors or controlled HDR protocols—limiting their applicability to typical 8-bit sRGB images from consumer devices. When applying 3DGS to LDR low-light scene reconstruction, two main issues arise: 1) unreliable SfM (Structure-from-Motion) initialization due to poor texture quality; and 2) noise overfitting during Gaussian optimization, resulting in degraded rendering results.

To tackle the aforementioned challenges, we propose LL-Gaussian, a novel framework for photorealistic low-light scene reconstruction and real-time pseudo normal-light novel view synthesis from standard sRGB images. First, we propose a end-to-end \textbf{L}ow-\textbf{L}ight \textbf{G}aussian \textbf{I}nitialization \textbf{M}odule (\textbf{LLGIM}) that leverages dense point cloud priors from learning-based MVS method and stochastic pruning with depth-guided refinement, enabling a reliable initialization process. Second, we introduce a novel dual-branch decomposition model that disentangles the scene into: Intrinsic Gaussian that capture intrinsic properties (reflectance and illumination), and Transient Gaussian that model unstable content (noise, color shifts, and illumination artifacts). This decomposition improves robustness and interpretability across views. Finally, we design an unsupervised optimization strategy incorporating physical constrains and diffusion prior to jointly guide decomposition and photorealistic enhancement. Our contributions are summarized as follows:

\begin{itemize}
    \item We propose LL-Gaussian, a novel framework for reconstructing LDR low-light scenes and synthesizing pseudo normal-light novel views from noisy sRGB inputs. Compared to NeRF-based methods, LL-Gaussian achieves up to $2000\times$ faster rendering speed and reduces training time to just $2\%$, while delivering superior rendering quality.
    \item We propose, for the first time, an end-to-end \textbf{L}ow-\textbf{L}ight \textbf{G}aussian \textbf{I}nitialization \textbf{M}odule (\textbf{LLGIM}) that generates robust initial point clouds without relying on conventional SfM approach, effectively addressing the initialization bottleneck under extreme low-light conditions.
    \item We design a dual-branch Gaussian decomposition model and a fully unsupervised optimization strategy that separates intrinsic scene attributes from transient degradation, improving robustness to noise and lighting artifacts during training.
    \item We contribute a challenging scene dataset collected in extreme low-light real-world environments, demonstrating the effectiveness of LL-Gaussian with extensive experiments.
\end{itemize}
\vspace{-3ex}

\begin{figure*}
    \centering
    \setlength{\abovecaptionskip}{0.2cm}
\includegraphics[width=\linewidth]{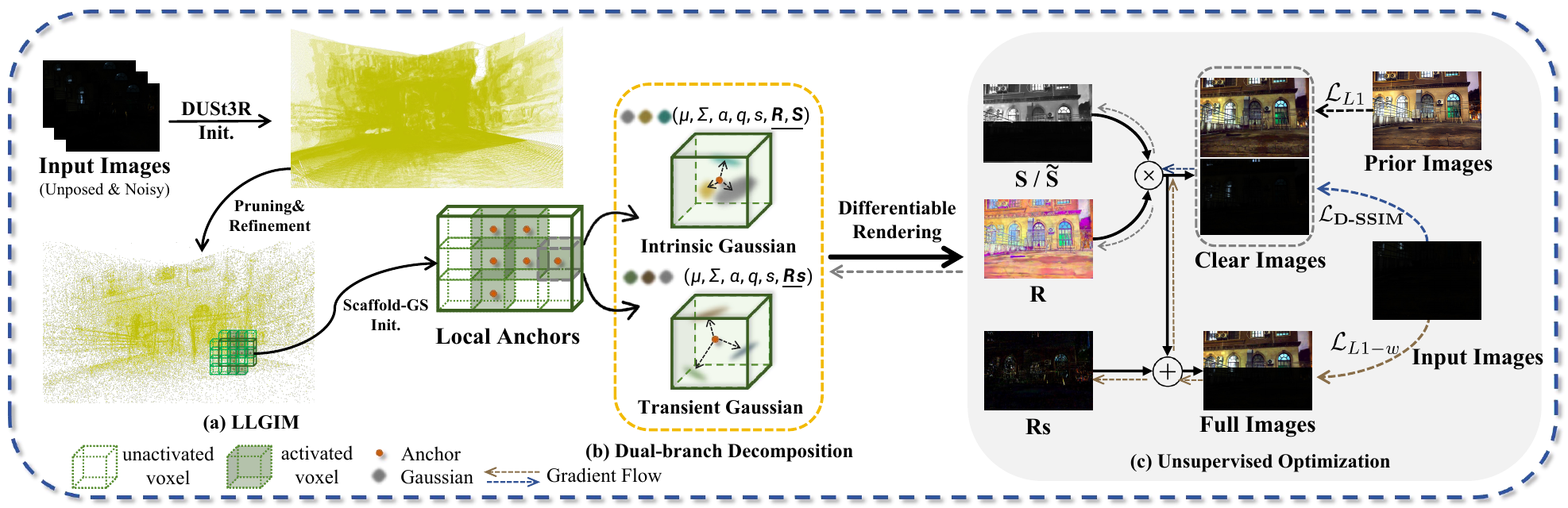}

\caption{Overview of the LL-Gaussian pipeline. (a) Given a set of unposed low-light images, our method first employs DUSt3R~\cite{wang2024dust3r} to generate dense point clouds, which are pruned and refined by the proposed LLGIM. (b) Initialized anchors are passed for Gaussian optimization, where a dual-branch decomposition is applied: the Intrinsic Gaussian branch captures static reflectance and illumination, while the Transient Gaussian branch models dynamic residuals. The decomposed Gaussians are rendered via differentiable splatting to component maps. (c) Unsupervised optimization leverages input and prior images to jointly optimize the Gaussian attributes and enhancement module. }
\label{fig:pipeline}
\vspace{-0.3cm}
\end{figure*}
\section{Related Work}

\noindent\textbf{Intrinsic Decomposition.}
Intrinsic decomposition separates visual content into reflectance and illumination components for better interpretability and downstream applications. In the 2D domain, both classical Retinex models~\cite{land1977retinex} and learning-based methods~\cite{wu2022uretinex,jiang2024lightendiffusion,retinexformer,yi2023diff,zhu2020zero} address this task from single images. In the 3D domain, recent neural rendering works~\cite{zhang2021nerfactor,yao2022neilf,ye2023intrinsicnerf,srinivasan2021nerv,jiang2023gaussianshader} incorporate intrinsic decomposition into volumetric or point-based 3D representations. However, these methods often assume well-lit, clean inputs. In contrast, we propose a decomposition strategy tailored for noisy, low-light sRGB inputs that separates stable intrinsic properties from transient degradations in a physically consistent manner.

\noindent\textbf{Learning-based Initialization for 3D Reconstruction.}
Conventional SfM pipelines like COLMAP~\cite{schonberger2016structure} often fail under low-light due to unreliable feature extraction. Learning-based MVS methods including MVSNet~\cite{yao2018mvsnet}, DUSt3R~\cite{wang2024dust3r} and Fast3R~\cite{yang2025fast3r} improve robustness via learned correspondence matching. Recent 3DGS approaches~\cite{fan2024instantsplat,hong2024pf3plat,liu2024mvsgaussian,chen2024mvsplat360} adopt MVS point clouds for Gaussian initialization, highlighting the value of geometry priors. However, all of these methods assume well-lit inputs and degrade under low-light conditions. To address this limitation,  we propose the first initialization module tailored for low-light scenes, leveraging DUSt3R’s robust MVS priors to produce compact, high-quality point clouds for efficient Gaussian optimization.

\noindent\textbf{Novel View Synthesis in Low-light Scenes.}
NeRF-based methods ~\cite{mildenhall2022nerf,wang2023lighting,cui2024aleth,zheng2024learning} enable novel view synthesis under low-light or HDR settings by leveraging implicit noise suppression and reflectance-illumination decomposition. However, their reliance on per-ray optimization and implicit MLPs leads to slow inference and biased convergence under severe noise. 3D Gaussian Splatting (3DGS)~\cite{kerbl20233d} offers real-time rendering with explicit scene representation, and recent extensions~\cite{li2024chaos, jin2024lighting, ye2024gaussian} adapt it to RAW or multi-exposure inputs. Yet, these approaches fail or perform suboptimally on LDR reconstruction due to lack of metadata and unstable initialization. We address these issues by introducing a robust initialization strategy and a dual-branch Gaussian decomposition model tailored for real-world low-light scenarios.

\vspace{-1ex}
\section{Preliminaries}
\label{sec:preliminaries}

\noindent\textbf{3D Gaussian Splatting (3DGS).} 
As an explicit scene representation paradigm, 3DGS~\cite{kerbl20233d} models 3D environments using a collection of anisotropic Gaussians that preserve differentiability while enabling real-time rendering. Each Gaussian primitive is parameterized by its position (mean $\mu \in \mathbb{R}^3$), covariance matrix $\Sigma \in \mathbb{R}^{3\times3}$, opacity $\alpha \in [0,1]$, and view-dependent color represented by spherical harmonics coefficients. The covariance matrix is decomposed into learnable scaling $S \in \mathbb{R}^3$ and rotation $R \in SO(3)$ parameters through $\Sigma = RSS^TR^T$. 

The rendering process employs a tile-based differentiable rasterizer that projects 3D Gaussians onto the image plane as 2D splats. For a pixel at position $x'$, the blended color $C(x')$ is computed via alpha compositing:
\begin{equation}
    C\left(x'\right) = \sum_{i \in N} T_i c_i \sigma_i, \quad \sigma_i = \alpha_i G_i'\left(x'\right),
\label{eq:gaussian splatting}
\end{equation}
where $G_i'$ denotes the projected 2D Gaussian, $T_i = \prod_{j=1}^{i-1}(1-\sigma_j)$ represents accumulated transmittance, and $N$ is the set of depth-ordered Gaussians overlapping the pixel.

\noindent\textbf{Scaffold-GS Architecture.} 
\label{sec:Scaffold-GS}
We build our method upon Scaffold-GS~\cite{lu2024scaffold}, which introduces structural regularization through anchor-based neural Gaussian generation. Each anchor at position $x_v$ emits $k$ neural Gaussians with positions determined by:
\begin{equation}
\left\{ \mu_0, \ldots, \mu_{k-1} \right\} = x_v + \left\{ \mathcal{O}_0, \ldots, \mathcal{O}_{k-1} \right\} \cdot l_v
\end{equation}
where $\mathcal{O}_i$ denotes predicted offsets and $l_v$ controls spatial distribution scale. Gaussian attributes including opacity $\alpha_i$, scale $S_i$, rotation $R_i$, and spherical harmonics coefficients are decoded from anchor features $\hat{\mathbf{f}}_v$ through lightweight MLPs conditioned on viewing parameters:
\begin{equation}
\left\{ \alpha_0, \ldots, \alpha_{k-1} \right\} = \mathrm{F}_\alpha(\hat{\mathbf{f}}_v, \Delta_{\mathrm{vc}}, \tilde{\mathbf{d}}_{\mathrm{vc}}),
\end{equation}
where $\Delta_{\mathrm{vc}}$ and $\tilde{\mathbf{d}}_{\mathrm{vc}}$ represent relative viewing distance and direction respectively. This scaffold structure enables efficient geometric regularization through anchor-level densification and pruning.

\section{Method}
In this paper, we introduce a novel method named LL-Gaussian, designed for normal-light novel view synthesis from degraded low-light multi-view sRGB images (8-bit per channel). Our approach is motivated by the following key challenges and solutions: 1) To tackle the difficulties of conventional SfM initialization under extreme low-light conditions, we propose the LLGIM (Sec. \ref{sec:LLGIM}). This module provides a robust Gaussian initialization process, enhancing the stability and accuracy of scene reconstruction. 2) To facilitate robust Gaussian optimization and controllable lighting manipulation, we introduce a dual-branch decomposition model (Sec. \ref{sec:decomposition and enhancement}). This model comprises the Intrinsic Gaussian, which models reflectance and illumination attributes, and the Transient Gaussian, which captures residual attributes to represent static scene properties and dynamic interference signals, respectively. 3) For accurate decomposition and high-quality enhancement, we design an unsupervised optimization strategy (Sec. \ref{sec:unsupervised optimization strategy}). This strategy integrates physical and diffusion priors to guide the optimization process effectively.
\vspace{-0.2cm}
\subsection{Low-Light Gaussian Initialization Module}
\label{sec:LLGIM}
To overcome the conventional Gaussian initialization limitations in low-light condition, we propose the \textbf{L}ow-\textbf{L}ight \textbf{G}aussian \textbf{I}nitialization \textbf{M}odule (\textbf{LLGIM}). The full implementation is detailed in Algorithm 1 of supplementary material.

\begin{figure}[t]
\centering
    \setlength{\abovecaptionskip}{0.2cm}
    \includegraphics[width=\linewidth]{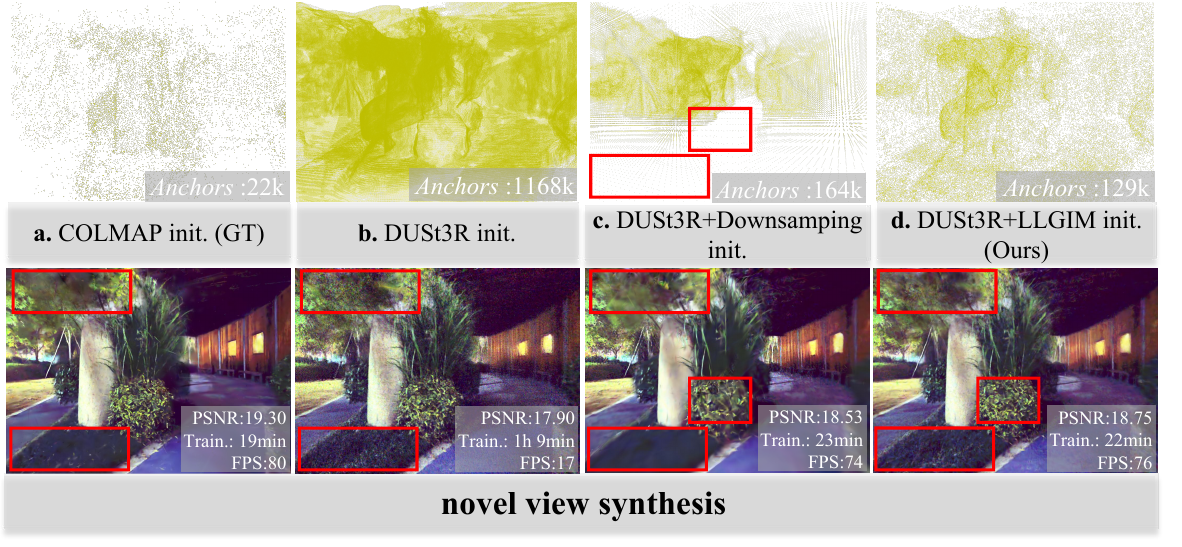}
   \caption{Ablation Studies on LLGIM (\textit{Zoom in for best view}). Note that (GT) denotes initialization with ground truth (normal-light) inputs, while the others using the low-light inputs for initialization.}
\label{fig:ablation1}
\vspace{-0.5cm}
\end{figure}

\noindent\textbf{Dense Point Cloud Prior Injection.} Our LLGIM module begins with dense point clouds from a learning-based MVS model (we adopt DUSt3R~\cite{wang2024dust3r}) that generates a set of densely covered and pixel-aligned point clouds with pairs of input images. While these pixel-wise 3D point clouds provide essential geometric cues for low-light scenes, direct Gaussian initialization with these overparameterized points introduces optimization inefficiencies, leads to artifacts and slow rendering speed, as illustrated in Fig. \ref{fig:ablation1}.

\noindent\textbf{Distance-Adaptive Stochastic Pruning.}
To address the redundancy issue in the dense point clouds reconstructed by DUSt3R while preserving essential geometric structures, we establish an adaptive probabilistic model that progressively filters redundant points through dynamic distance constraints.

Given the input point cloud $\mathcal{P}=\{x_i\}^N_{i=1}$, we first construct a voxel grid $\mathcal{V}$ with resolution $r$ following the scene parameterization approach of Scaffold-GS~\cite{lu2024scaffold}. Each voxel $v_j \in \mathcal{V}$ aggregates points within its spatial domain, generating candidate anchors (structured manager of gaussian primitives) through spatial computation. This spatial discretization naturally induces local density awareness while maintaining structural continuity.

For each candidate anchor $a_k \in \mathcal{A}$, we define its preservation probability through an energy-based formulation:
\begin{equation}
    P(a_k)=\min (1, \frac{d_{min}(a_k)}{\tau^{(t)}}+\epsilon),
\end{equation}

where $d_{min}(a_k)$ quantifies the minimum inter-anchor distance at iteration $t$, $\tau^{(t)}$ is an adaptive distance threshold, and $\epsilon$ ensures numerical stability. The filtering process adopts a stochastic Bernoulli sampling governed by $P(a_k)$, implementing soft suppression of redundant anchors while preserving structural critical points with probabilistic guarantees.

To achieve progressive refinement, we design a threshold update rule: 
\begin{equation}
    \tau^{(t+1)}=\tau^{(t)}\cdot \exp(\beta \cdot \frac{|\mathcal{A}^{(t)}|}{|\mathcal{A}^{(0)}|})
\end{equation}
where $\beta$ is a temperature parameter and $|\mathcal{A}^{(t)}|$ represents the current anchor count. This annealing strategy enables iterative processing: Early iterations with smaller $\tau$ focus on removing obvious redundancies, while subsequent stages with increased $\tau$ relax spatial constraints, thus reducing potential redundancies.

\noindent\textbf{Depth-Guided Warm-up Refinement.}
To further address the persistent geometric artifacts (e.g., floaters, distorted surfaces) in pruned point clouds while recovering valid structures over-filtered during stochastic pruning, we introduce a depth-guided warm-up refinement. The key lies in the synergistic integration of monocular depth prior distillation and progressive geometric rectification, where a pre-trained monocular depth estimator serves as both artifact detector and geometric corrector through differentiable optimization. To provide scale-invariant supervision, we utilize  PCC-based (Pearson Correlation Coefficient) loss to measure linear dependence between the rendered depth $\hat{D}_k$ and prior depth $D_k^{mono}$ which is provided by Depth Anything V2~\cite{yang2024depth}:
\begin{equation}
\mathcal{L}_{depth} = 1 - \frac{\text{Cov}(\hat{D}_k, D_k^{mono})}{\sigma\{\hat{D}_k\} \sigma\{D_k^{mono}\}}
\end{equation}
where $\text{Cov}(\cdot,\cdot)$ denotes covariance and $\alpha\{\cdot\}$ represents standard deviation.

\subsection{3D Scene Decomposition Modeling with Dual Gaussian Primitives}
\label{sec:decomposition and enhancement}
To address the challenges of novel view synthesis under low-light conditions with transient interference, we propose a dual-branch Gaussian decomposition model that explicitly disentangles the scene into intrinsic attributes and transient disturbances. Our model comprises two specialized Gaussian sets: \textit{Intrinsic Gaussian}  and \textit{Transient Gaussian}, as shown in Fig. \ref{fig:pipeline}.

\begin{figure}[t]
\centering
    \setlength{\abovecaptionskip}{0.2cm}
    \includegraphics[width=\linewidth]{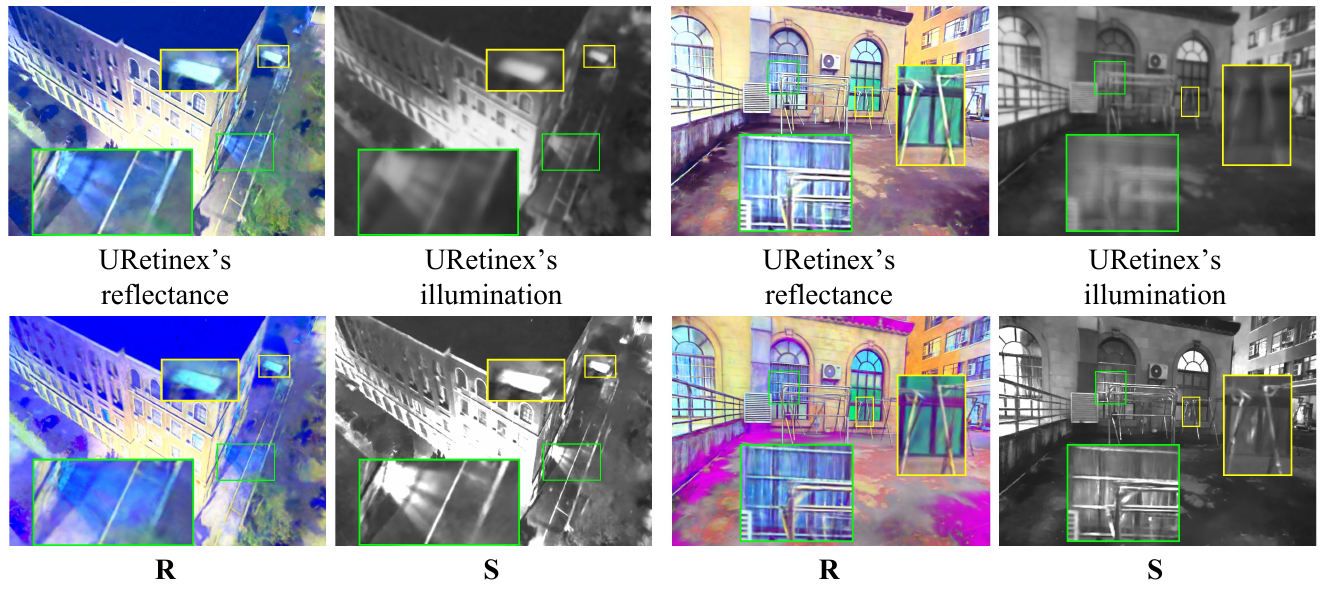}
   \caption{Visualization of our intrinsic decomposition results (the second row) compared with the image-based decomposition model URetinex~\cite{wu2022uretinex}. Our method successfully disentangles material-dependent properties from illumination effects. Note that \textbf{S} is brightened for a better view.}

\label{fig:decomposition_result}
\vspace{-0.5cm}
\end{figure}

\noindent\textbf{Intrinsic Gaussian for Static Decomposition.} 
The Intrinsic Gaussian, denoted as $\mathcal{G}_d=\{{g_p^d}\}_{p=1}^{M_d}$, models the inherent static scene properties through a physically-inspired decomposition into reflectance and illumination components. Each Gaussian primitive $g_p^d$ is parameterized by $(\mu_p^d, \Sigma_p^d,\alpha_p^d,R_p,S_p)$, where $R_p \in \mathbb{R}^3$ represents the surface reflectance, and $S_p \in \mathbb{R}^+$ denotes the illumination intensity. Building upon the Scaffold-GS framework (see Sec.~\ref{sec:Scaffold-GS}), each Intrinsic Gaussian's attributes are decoded from its corresponding anchor feature. To disentangle the static scene properties into physically meaningful components, we design two tiny MLPs: $\mathit{F}_{R}$ for reflectance and $\mathit{F}_{S}$ for illumination. These MLPs operate atop the scaffold’s shared feature volume, ensuring spatial consistency while maintaining computational efficiency. Specifically, for each anchor feature $\hat{\mathbf{f}}_v^d$ , the components are decoded as:
\begin{equation}
    \left\{ R_0, \dots, R_{k-1} \right\} = \mathit{F}_{R}(\hat{\mathbf{f}}_v^d, \Delta_{vc}),
\end{equation}
\begin{equation}
    \left\{ S_0, \dots, S_{k-1} \right\} = \mathit{F}_{S}(\hat{\mathbf{f}}_v^d, \Delta_{vc},\tilde{d}_{vc}),
\end{equation}
where $\Delta_{\mathrm{vc}}$ denotes the relative viewing distance between the anchor center and the camera, and $\tilde{d}_{vc}$ represents the normalized view direction. These geometry-aware features explicitly encode spatial and view-dependent relationships, enabling the MLPs to disentangle illumination-invariant reflectance from illumination intensity, as illustrated in the second row of Fig. \ref{fig:decomposition_result}. The lightweight architecture of $\mathit{F}_{R}$ and $\mathit{F}_{S}$ (with 1 hidden layer) ensures efficient feature specialization while preventing overfitting. 

\noindent\textbf{Transient Gaussian for Dynamic Residual Modeling.} 
The transient Gaussian, denoted as $\mathcal{G}_r=\{g_q^r\}_{q=1}^{M_r}$, models dynamic interference. Each Gaussian primitive $g_q^r$ is parameterized by $(\mu_q^r,\Sigma_q^r,\alpha_q^r,Rs_q)$, where $Rs_q \in \mathbb{R}^3$ captures transient residual attributes (e.g., sensor noise or transient illumination artifacts). $\mathcal{G}_r$ shares the voxel-grid anchor structure with $\mathcal{G}_d$ to ensure spatial consistency, but maintains \textbf{independent anchor feature} $\hat{\mathbf{f}}_v^r$ for disentangling transient effects. Inspired by ~\cite{martin2021nerf, kulhanek2024wildgaussians, lin2024vastgaussian}, we additionally introduce a per-image learnable embedding $\textbf{e}_j \in \mathbb{R}^{r_e}$ that captures transient variations specific to the $j$-th input view. Although the anchor positions and geometry-aware terms ($\Delta_{vc}, \tilde{d}_{vc}$) are inherited from the shared scaffold, $\mathcal{G}_r$ employs a separate feature volume and per-image embedding $\textbf{e}_j$ to isolate transient properties from static scene attributes.

To decode the residual component, we design a tiny MLP $\mathit{F}_{Rs}$ conditioned on transient-specific features and per-view variations:
\begin{equation}
\{ Rs_0, \dots, Rs_{k-1} \} = \mathit{F}_{Rs}\left(\hat{\mathbf{f}}_v^r, \Delta_{vc}, \tilde{d}_{vc}, \textbf{e}_j\right).
\end{equation}

Note that since the transient branch is only active during training, the per-image embedding $\textbf{e}_j$ is only required for encoding training views.

\noindent\textbf{Differentiable Rendering.}
The Intrinsic and Transient Gaussian is rendered into pixel-aligned maps through differentiable splatting (cf. Eq.~\ref{eq:gaussian splatting}). For each view, the Intrinsic Gaussian $\mathcal{G}_d$ is aggregated to render the reflectance map $\mathbf{R}$ and the illumination map $\mathbf{S}$, while the Transient Gaussian $\mathcal{G}_r$ is splatted on the residual map $\mathbf{Rs}$.

The final pixel color $\mathbf{\hat{C}}_{low}$ is synthesized by coupling the intrinsic photometric components with the transient residual:
\begin{equation}
    \mathbf{\hat{C}}_{low} = \underbrace{\mathbf{R} \odot \mathbf{S}}_{\text{Intrinsic components}} + \underbrace{\mathbf{Rs}}_{\text{Transient residual}},
\end{equation}
where $\odot$ denotes element-wise multiplication.

\noindent\textbf{Illumination Enhancement.}
Building upon the decomposed illumination attributes $\{S_p\}$ from Intrinsic Gaussian, we propose a neural tone-mapping module $\mathcal{T}$ (a tiny MLP with 1 hidden layer) that adaptively enhances low-light conditions while preserving physical plausibility. By rendering as Eq.~\ref{eq:gaussian splatting}, the enhanced illumination map $\tilde{\mathbf{S}}\in \mathbb{R}^3$ can be rendered by:

\begin{equation}
    \tilde{\mathbf{S}} = \sum_{p \in N}T_p\sigma_p\mathcal{T}(S_p,\mathbf{\hat{f}}^d_v; \theta),
\label{eq:enhanced illumination map}
\end{equation}
where $\theta$ denotes learnable parameters. Note that 3-channel $\tilde{\mathbf{S}}$ is designed for correcting color bias during enhancement stage. 

The enhancement preserves reflectance consistency, ensuring surface material authenticity. The enhanced pixel color is then computed as:
\begin{equation}
    \mathbf{\hat{C}}_{nor}=\mathbf{R}\odot \tilde{\mathbf{S}}.
\label{eq:C_nor}
\end{equation}

\subsection{Unsupervised Optimization Strategy}
\label{sec:unsupervised optimization strategy}
We design a suite of unsupervised losses to jointly optimize the Gaussian attributes and enhancement module $\mathcal{T}$:

\noindent\textbf{Reconstruction loss.} Following the original Scaffold-GS~\cite{lu2024scaffold} framework, we adopt L1 loss combining with DSSIM loss for pixel-wise fidelity and structural consistency. To address the critical challenge of recovering subtle details in low-intensity regions, we introduce the weighted L1 loss $\mathcal{L}_{L1-w}$ which is inspired by RawNeRF~\cite{mildenhall2022nerf}, as:
\begin{equation}
    \mathcal{L}_{L1-w}=\left\|\frac{\mathbf{\hat{C}}_{low}-\mathbf{C}_{low}}{\mathbf{sg}(\mathbf{\hat{C}}_{low})+\epsilon}\right\|_1,  
\end{equation}
where $\mathbf{\hat{C}}_{low}=\mathbf{R}\odot\mathbf{S}+\mathbf{Rs}$ denotes the predicted results, $\mathbf{C}_{low}$ is the low-light input, and the $\mathbf{sg(\cdot)}$ represents the stop-gradient operation.

To further decouple static and transient components during optimization, we let DSSIM loss optimize only the intrinsic attributes, exploiting its structural robustness to prevent transient residuals from corrupting static scene reconstruction. The reconstruction loss can be written as:
\begin{equation}
\begin{aligned}
    \mathcal{L}_{recon}=&(1-\lambda)\mathcal{L}_{L1-w}(\mathbf{\hat{C}}_{low},\mathbf{C}_{low})\\
&+\lambda\mathcal{L}_{DSSIM}(\mathbf{\hat{C}}_{low}-\mathbf{Rs},\mathbf{C}_{low})
\end{aligned}
\end{equation}

\noindent\textbf{Illumination Prior.} Natural illumination exhibits local smoothness in textural regions and sharpness along structural edges. We formulate smooth prior loss as:
\begin{equation}
    \mathcal{L}_{smo} = \| w_x \cdot\partial_x \textbf{S} \|_1 + \| w_y \cdot\partial_y \textbf{S} \|_1,
\end{equation}
\begin{equation}
    w_x=\frac{1}{\partial_x (\mathcal{G}_{lp}\circ \mathbf{C}_{g}) +\epsilon},
    w_y=\frac{1}{\partial_y (\mathcal{G}_{lp}\circ \mathbf{C}_{g})+\epsilon},
\end{equation}
where $\mathcal{G}_{lp}$ is a Gaussian low-pass filter, $\circ$ denotes the convolution operator, $\mathbf{C}_{g}$ represents the gray-scale low-light input images, and $\epsilon$ prevents division by zero. This adaptively relaxes the smoothness constraints near intensity edges. Inspired by~\cite{guo2016lime}, we initialize illumination estimation with the maximum chromaticity, providing a coarse guidance for illumination disentanglement during early training phases as $\mathcal{L}_{init} = \| \mathbf{S} - \max_{u \in \{R,G,B\}} \mathbf{C}_{low}^u \|_1,$ the joint illumination prior loss becomes:
\begin{equation}
\mathcal{L}{_{ill}} =\mathcal{L}_{init}+\lambda_{smo}\mathcal{L}_{smo}.
\end{equation}
where $\lambda_{smo}$ is set to 0.001.

\noindent\textbf{Residual Constraint.} To prevent residual components overfit static information in early optimization, We leverage a regular L1 loss:
\begin{equation}
    \mathcal{L}_{re}=\lambda_{re}\|\mathbf{Re}\|_1 
\end{equation}
The weight $\lambda_{re}$ is set higher in early optimization to let $\mathbf{R}\odot \mathbf{S}$ close to the target  $\tilde{\mathbf{C}}_{low}$ and dropped gradually.

\noindent\textbf{Enhancement Supervision.} To establish photorealistic consistency between enhanced novel views and physical constraints, we formulate a dual-constrained regularization framework for optimizing the neural tone-mapping module $\mathcal{T}$:
\begin{equation}
\begin{aligned}
\mathcal{L}_{enh} &= \left\|\frac{\tilde{\mathbf{S}}}{\mathbf{sg}(\mathbf{S}) + \epsilon} - \gamma  \right\|_1+
\left\| \mathbf{\hat{C}}_{nor}- \mathbf{C}_{pri}\right\|_1,
\label{eq:enhancement}
\end{aligned}
\end{equation}
where $\gamma$ denotes the user-specified enhancement ratio, $\tilde{{\mathbf{S}}}$ is the enhanced illumination map rendered by Eq.~\ref{eq:enhanced illumination map} and $\mathbf{\hat{C}}_{nor}$ is computed by Eq. \ref{eq:C_nor}. The first regularization term enforces parametric control over enhancement intensity through the coefficient $\gamma$. while the second term introduces learned color constancy priors via $\mathbf{C}_{pri}=\mathcal{D}^*(\gamma\cdot \mathbf{C}_{low};\theta')$, where 
$\mathcal{D}^*$ denotes a frozen pre-trained diffusion model (we employ a image restoration model, StableSR ~\cite{wang2024exploiting}) that establishes data-driven color prior through its hierarchical denoising architecture. Notably, this formulation circumvents the need for explicit illumination estimation and chromaticity modeling required by conventional physical priors. Instead, it capitalizes on the diffusion model's implicit understanding of natural color distributions acquired through large-scale visual data training, thereby ensuring photometrically plausible enhancements while maintaining scene-adaptive color fidelity.

\begin{figure*}
\centering
    \setlength{\abovecaptionskip}{0.2cm}
\includegraphics[width=\linewidth]{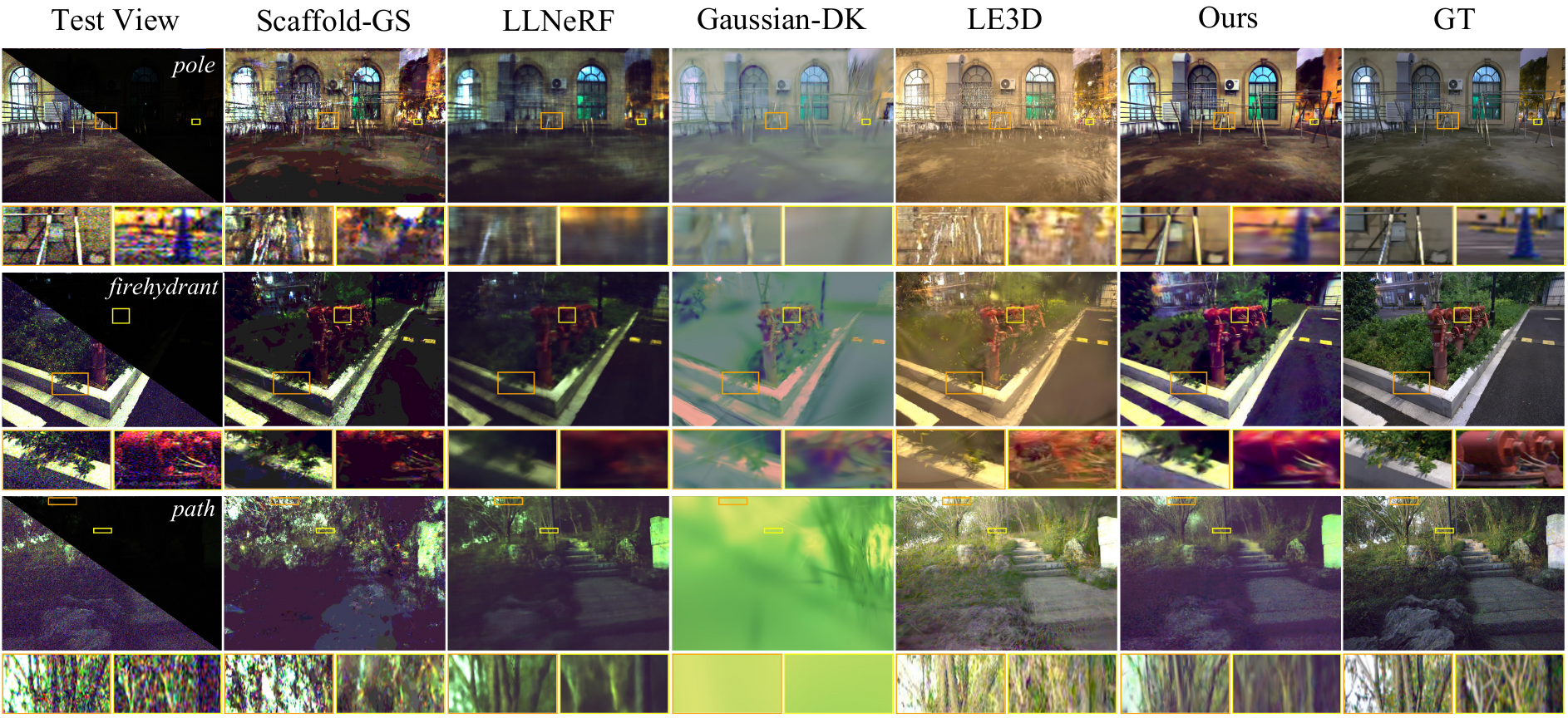}
\caption{Visualization comparison of novel view synthesis results of LL-Gaussian (Ours) and other baseline methods on our LLRS Dataset. Note that test input views are brightened for a better view.}
\label{fig:results1}
\end{figure*}

Above all, LL-Gaussian's loss function $\mathcal{L}$ includes four components: 3DGS photometric loss $\mathcal{L}_{recon}$, unsupervised prior constraints $\mathcal{L}_{ill}$, residual regularization $\mathcal{L}_{re}$ and enhancement loss $\mathcal{L}_{enh}$. The final training loss is:
\begin{equation}
    \mathcal{L}=\mathcal{L}_{recon} + \lambda_{ill}\mathcal{L}_{ill} + \lambda_{re}\|\mathbf{Re}\|_1 +\lambda_{enh}\mathcal{L}_{enh}
\end{equation}
where $\lambda_{ill}=1$. We set  $\lambda_{re}=2$ in first iteration and dropped to 0.5 in the later iterations, while $\lambda_{enh}=0$ in first 2k iterations and then set to 1.0 in later iterations.

\vspace{-1ex}
\section{Experiments}
\subsection{Challenging Real-world Multi-view Dataset}
\label{sec:dataset}

Existing datasets for novel view synthesis under challenging low dynamic range (LDR) low-light conditions remain limited in both scale and realism. Although LLNeRF~\cite{wang2023lighting} is an early effort that captures 12 real-world low-light scenes, it lacks corresponding normal-light reference images, which hinders comprehensive evaluation. Aleth-NeRF~\cite{cui2024aleth} further advances the field by introducing the LOM dataset, consisting of five scenes captured under multi-illumination (low-light, normal-light, and over-exposed). However, its practical is constrained by the small scale of the scenes. More importantly, both datasets adopt a simplified forward-facing captures result in bounded views with limited diversity—creating a domain gap from real-world settings.

To address these issues, we introduce \textbf{LLRS}, an multi-view dataset featuring 8 \textbf{L}ow-\textbf{L}ight \textbf{R}eal-world \textbf{S}cenes (6 captured handheld / 2 with a UAV). Each scene contains 25–45 images ($1024 \times 768$) of unbounded outdoor environments recorded with a Canon EOS R8 and DJI Mini 3. To emulate real-world degradations, we use adaptive exposure (10–250 ms) and ISO (3200–12800), and ensure: 1) Diverse viewpoints: Unlike forward-facing setups, we varied shooting angles and distances to ensure subjects in dark scenes remain visible. As a result, LLRS includes a mix of wide-angle and close-up shots; 2) Realistic low-light effects: sRGB images are generated from RAW using standard ISP without extra post-processing (e.g., denoising), preserving natural degradations such as motion blur and heavy noise—especially in ultra-dark regions; 3) Rich lighting diversity: Scenes span moonlit (0.1 lux) to streetlight-interrupted darkness, covering a wide range of real-world illumination scenarios.

\begin{figure*}
\centering
    \setlength{\abovecaptionskip}{0.2cm}
\includegraphics[width=\linewidth]{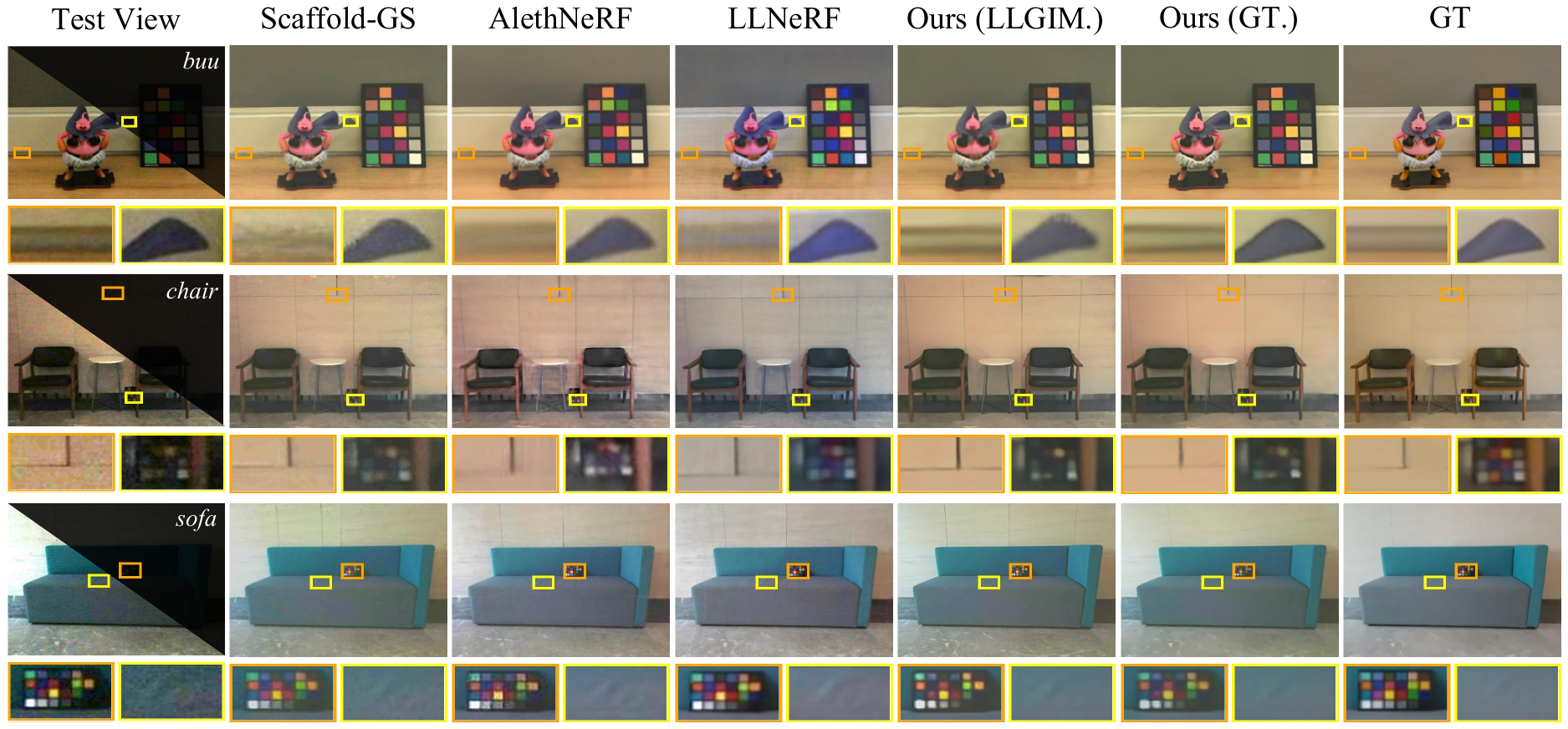}
\caption{Visualization comparison of novel view synthesis results of LL-Gaussian (Ours) and other baseline methods on LOM Dataset. Note that test input views are brightened for a better view.}
\label{fig:results2}
\end{figure*}

\vspace{-0.2cm}
\subsection{Experiment Setup}
\noindent\textbf{Datasets.} To comprehensively evaluate our LL-Gaussian, we conduct experiments on three low-light scene datasets: \textbf{LOM}~\cite{cui2024aleth},dataset, \textbf{LLNeRF}~\cite{wang2023lighting} dataset and our \textbf{LLRS} dataset, detailed in Sec.~\ref{sec:dataset}. Note that results on \textbf{LLNeRF} dataset are presented in Sec. 6 of the supplementary material.

\noindent\textbf{Baselines.} We compare our method with state-of-the-art NeRF- and 3DGS-based approaches. On LLRS, we evaluate against AlethNeRF~\cite{cui2024aleth} (LDR), LLNeRF~\cite{wang2023lighting} (LDR), Gaussian-DK~\cite{ye2024gaussian} (HDR), LE3D~\cite{wang2023lighting} (HDR), and Scaffold-GS~\cite{lu2024scaffold} (LDR). On the LOM~\cite{cui2024aleth} and LLNeRF~\cite{wang2023lighting} datasets, which lack HDR supervision, we compare with the corresponding LDR methods.  Additionally, we include comparisons with LLIE (Low-Light Image Enhancement) + Scaffold baselines, reported in Sec. 5 of the supplementary material.

\noindent\textbf{Metrics.}  We follow common practice and employ PSNR, SSIM~\cite{wang2004image}, and LPIPS~\cite{zhang2018unreasonable} for our evaluation. SSIM and LPIPS are prioritized for their robustness to brightness variations. For comprehensive performance characterization, we also report training times in GPU hours of an NVIDIA Tesla V100 as well as rendering times in frames-per-second (FPS).

\vspace{-0.2cm}
\subsection{Results Analysis}

\noindent\textbf{Comparison on our LLRS Dataset.}
Due to the failure of COLMAP initialization under extreme low-light conditions, the baseline inputs are generated using camera poses and initial point clouds derived from COLMAP initialization on paired ground-truth (GT) images. In contrast, our method directly uses DUSt3R-estimated point clouds and camera poses extracted from the low-light inputs themselves. As shown in Fig. \ref{fig:results1} and Table \ref{tab:results1}, our approach significantly outperforms both NeRF- and 3DGS-based baselines.For NeRF-based methods:
1) LLNeRF struggles with noise suppression and color correction due to its implicit representation, leading to blurry textures and color shifts; 2) AlethNeRF fails to converge across all scenes, likely due to sparse viewpoints and weak photometric supervision under extreme low-light. In contrast, our method preserves fine details and accurate colors, thanks to explicit residual modeling and diffusion-based color prior regularization. Among 3DGS-based methods: 1) Scaffold-GS suffers from geometric artifacts due to its noise-sensitive density estimation; 2) Gaussian-DK exhibits color distortions as it relies on unavailable multi-exposure inputs; 3) LE3D recovers reasonable colors via sensor metadata but produces suboptimal geometry. In contrast, our method achieves robust optimization against noise and reliable color reconstruction. Notably, it offers a \textbf{700$\times$} speedup in training and \textbf{2000$\times$} faster real-time rendering compared to NeRF-based approaches.

\noindent\textbf{Comparison on the LOM Dataset.}
To validate the generalizability of our approach, we conduct comprehensive evaluations on the AlethNeRF dataset comprising calibrated sRGB images of small-scale indoor scenes. For initial point clouds and camera poses input, we adopted a similar configuration while additionally incorporating the comparative results of our method under ground-truth (GT) initialization.
As demonstrated in Fig. \ref{fig:results2} and Table \ref{tab:results2}, our method outperforms all baselines in both visual quality and quantitative metrics. Notably,  the results obtained by LLGIM initialization strategy attains comparable reconstruction accuracy to GT COLMAP initialization, and exhibits superior computational efficiency. Compared to NeRF-based methods, our method requires only \textbf{2\%} of the training time and achieves \textbf{500$\times$} rendering speed improvement. These results confirm our method's adaptability across indoor scene types in low-light imaging conditions.


\begin{figure*}
\centering          
\setlength{\abovecaptionskip}{0.1cm}
\includegraphics[width=\linewidth]{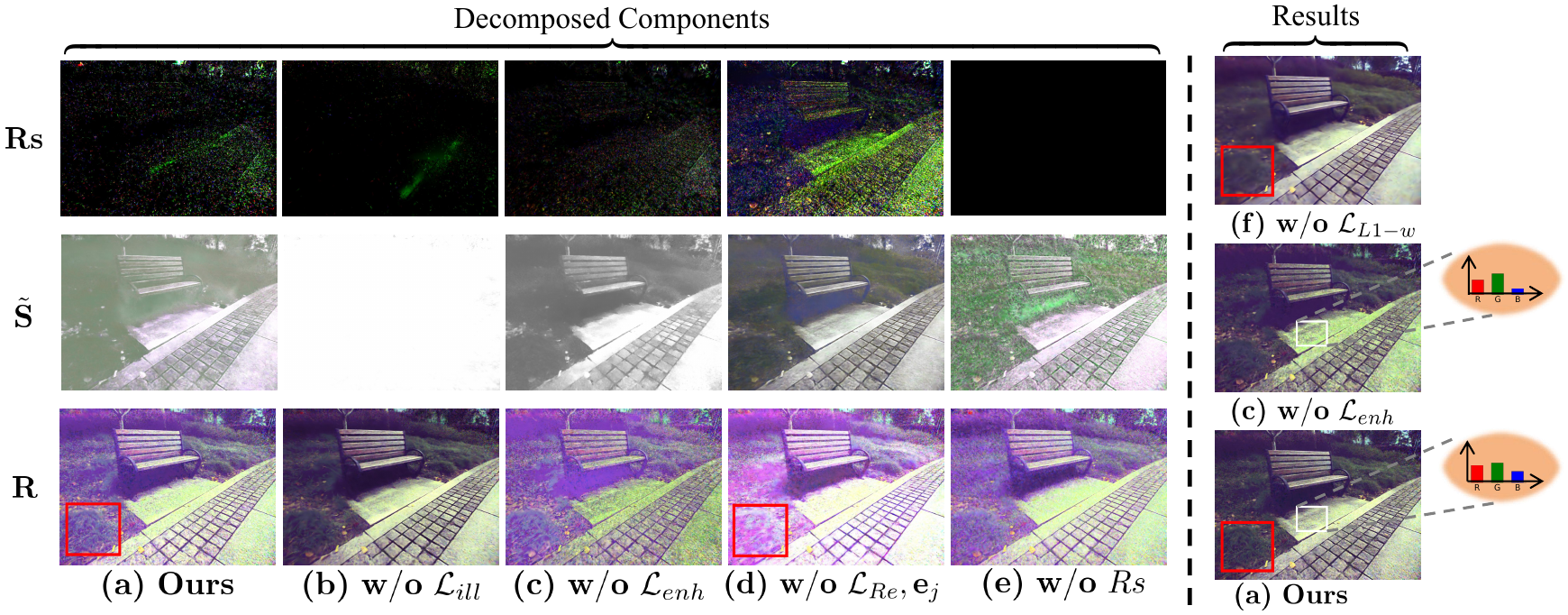}
\caption{Ablation Studies of Unsupervised Optimization Strategy \& Residual Modeling (\textit{Zoom in for best view}).}
\label{fig:ablation2}
\end{figure*}

\begin{table*}[tb]
\centering
\setlength{\abovecaptionskip}{0.2cm}
\caption{Quantitative comparisons with baseline methods on our proposed dataset. We report three rendering quality metrics (SSIM, LPIPS, PSNR) and two efficiency metrics (FPS and training times(GPU·h)). The \colorbox{red!30}{first},  \colorbox{orange!30}{second} and \colorbox{yellow!30}{third} best-performing methods highlighted. Our method shows overall superior performance over state-of-the-art baseline methods and offer real-time rendering.}
\renewcommand\arraystretch{0.9}
\setlength{\tabcolsep}{1.2pt}
{\footnotesize
\begin{tabular}{c c c ccc ccc ccc ccc ccc ccc}
\toprule
\multirow{2}{*}{\textbf{Methods}} & \multirow{2}{*}{\textbf{FPS}$\uparrow$}& \multirow{2}{*}{\textbf{Train.}$\downarrow$} & \multicolumn{3}{c}{\textbf{chair}} & \multicolumn{3}{c}{\textbf{firehydrant}} & \multicolumn{3}{c}{\textbf{path}} & \multicolumn{3}{c}{\textbf{pole}} & \multicolumn{3}{c}{\textbf{staircase}} & \multicolumn{3}{c}{\textbf{stone}}\\
\cline{4-21}
\rule{0pt}{10pt}
 & &  & SSIM$\uparrow$ & LPIPS$\downarrow$ & PSNR$\uparrow$ & SSIM$\uparrow$ & LPIPS$\downarrow$ & PSNR$\uparrow$ & SSIM$\uparrow$ & LPIPS$\downarrow$ & PSNR$\uparrow$ & SSIM$\uparrow$ & LPIPS$\downarrow$ & PSNR$\uparrow$ & SSIM$\uparrow$ & LPIPS$\downarrow$ & PSNR$\uparrow$ & SSIM$\uparrow$ & LPIPS$\downarrow$ & PSNR$\uparrow$\\
\midrule
Scaffold-GS & \best{96} & \best{0.27} & \third{0.208} & \third{0.699} & \second{16.27} & \third{0.382} & \second{0.658} & \second{17.93} & \third{0.166} & \third{0.692} & 15.25 & 0.439 & 0.627 & 18.48 & \second{0.292} & \second{0.667} & \second{15.49} & \third{0.275} & \third{0.656} & \second{17.32} \\
LLNeRF  & 0.03 & 16.93 & 0.118 & 0.778 & \third{15.84} & 0.233 & 0.727 & \third{17.62} & 0.157 & 0.722 & \third{17.53} & \third{0.454} & 0.649 & \third{19.87} & 0.163 & 0.689 & \third{14.21} & 0.167 & 0.731 & \third{17.04} \\
Gaussian-DK & \third{64} & \second{0.33} & \second{0.272} & 0.789 & 14.98 & \second{0.424} & \third{0.665} & 17.55 & - & - & - & \second{0.683} & \second{0.489} & \best{20.69} & \third{0.259} & 0.865 & 11.30  & 0.265 & 0.852 & 12.84 \\
LE3D        & 62 & 0.43 & 0.165 & \second{0.682} & 13.75 & 0.339 & 0.681 & 15.02 & \second{0.244} & \best{0.616} & \second{18.18} & 0.415 & \third{0.577} & 16.42 & 0.228 & \third{0.680}  & 12.89 & \second{0.286} & \second{0.632} & 16.84 \\
Ours        & \second{72} & \third{0.37} & \best{0.296} & \best{0.633} & \best{18.22} & \best{0.448} & \best{0.623} & \best{18.35} & \best{0.291} & \second{0.635} & \best{18.46} & \best{0.705} & \best{0.440}  & \second{20.25} & \best{0.378} & \best{0.621} & \best{15.71} & \best{0.368}  & \best{0.620} & \best{18.75} \\
\bottomrule
\end{tabular}
}
\label{tab:results1}
\vspace{-0.3cm}
\end{table*}


\begin{table}[tb]
\centering
\setlength{\abovecaptionskip}{0.2cm}
\caption{Quantitative comparisons with baseline methods on \textbf{LOM} dataset.  The \colorbox{red!30}{first},  \colorbox{orange!30}{second} and \colorbox{yellow!30}{third} best-performing methods highlighted. We significantly outperform all baseline methods.}

\resizebox{0.4\textwidth}{!}{\begin{tabular}{c c c ccc }
\toprule
\multirow{2}{*}{\textbf{Methods}} & \multirow{2}{*}{\textbf{FPS}$\uparrow$}& \multirow{2}{*}{\textbf{Train.}$\downarrow$} & \multicolumn{3}{c}{\textbf{LOM Dataset}} \\
\cline{4-6}
\rule{0pt}{10pt}
 & &  & SSIM$\uparrow$ & LPIPS$\downarrow$ & PSNR$\uparrow$ \\
\midrule
Scaffold-GS & \best{102}   & \second{0.23}  & 0.801 & 0.334 & \best{24.64} \\
LLNeRF      & 0.120 & 10.67 & \third{0.823} & \third{0.309} & 23.37 \\
AlethNeRF   & 0.109 & 9.31  & 0.792 & 0.316 & 23.85 \\
Ours (LLGIM.) & \second{70}    & \best{0.19}  & \second{0.839} & \best{0.217} & \second{24.52} \\
Ours (GT.)    &\third{50}    & \third{0.27}  & \best{0.845} & \second{0.219} & \third{24.51} \\
\bottomrule
\end{tabular}
}
\label{tab:results2}
\vspace{-12pt}
\end{table}

\vspace{-0.3cm}
\subsection{Ablation Study \& Analysis}
\noindent\textbf{LLGIM Initialization.} In our extreme low-light dataset, direct COLMAP initialization fails to obtain valid results. To validate the efficacy of the LLGIM module, we conduct comparative analyses with: (\textbf{a}) COLMAP init. using GT data, (\textbf{b}) DUSt3R init. and (\textbf{c}) DUSt3R+Downsampling init., as illustrated in Fig.~\ref{fig:decomposition_result}. It can be observed from Fig. \ref{fig:decomposition_result} (\textbf{b}) that LL-Gaussian without LLGIM tends to prolonged training times and slow rendering speed due to the huge redundant point clouds. Compared with Fig. \ref{fig:decomposition_result} (\textbf{c}), LLGIM achieves more accurate and detailed novel view synthesis. Notably, compared to with COLMAP init. (GT), LL-Guassian with LLGIM init. demonstrates competitive performance in novel view synthesis, both in reconstruction quality and efficiency. In some weakly textured regions, our method even achieves superior results.

\noindent\textbf{Intrinsic \& Transient Components.} To assess our intrinsic decomposition, we compare with URetinex~\cite{wu2022uretinex}, a representative image-based method. As shown in Fig.\ref{fig:decomposition_result}, our method effectively separates material-dependent properties (e.g., texture and albedo) from illumination, preserving specular highlights (e.g., on metal and glass) within the \textbf{S}. In contrast, URetinex exhibits residual lighting in the reflectance map, especially in high-reflectivity regions where specular patterns are misinterpreted as texture. Additionally, our illumination maps \textbf{S} retain sharp transitions at object boundaries and geometric discontinuities, whereas URetinex tends to over-smooth these regions. These results highlight the advantage of our approach in accurately disentangling view-dependent illumination from intrinsic appearance. For the transient component $Rs$, Fig. \ref{fig:ablation2}(e) reveals that the absence of $Rs$ induces degradation in both the illumination and reflectance estimations. Specifically, thin noise Gaussians emerge during decomposition, severely compromising the accuracy of intrinsic component separation.


\begin{table}[tb]
\centering
\setlength{\abovecaptionskip}{0.2cm}
\caption{Ablation Studies of Each Module on the our LLRS dataset.}
\resizebox{0.3\textwidth}{!}{\begin{tabular}{c  ccc }
\toprule
\textbf{Methods} &  SSIM$\uparrow$ & LPIPS$\downarrow$ & PSNR$\uparrow$ \\
\midrule
w/o LLGIM     & 0.274 & 0.607 & 17.12 \\
w/o Residual  & 0.334 & 0.635 & 17.59 \\
w/o all prior & 0.323 & 0.615 & 16.81 \\
Ours           & \textbf{0.414} & \textbf{0.595} & \textbf{18.29} \\
\bottomrule
\end{tabular}
}
\label{tab:ablation}
\vspace{-12pt}
\end{table}


\noindent\textbf{Unsupervised Optimization Strategy.} To evaluate the effectiveness of our unsupervised optimization strategy, we performed ablation studies on the proposed prior loss functions. As shown in Fig. ~\ref{fig:ablation2}(b,d), removing either the illumination prior or residual constraint leads to significant degradation in decomposition results. Fig. ~\ref{fig:ablation2}(c,f) further demonstrate that: (1) omitting the $\mathcal{L}_{L1-w}$ loss deteriorates detail reconstruction in rendering results; (2) removing the physical and diffusion priors adversely affects color correction performance. The quantitative results are shown in Table \ref{tab:ablation}.
\vspace{-0.2cm}
\subsection{Limitations}
Our method achieves high-quality, real-time reconstruction of challenging low-light scenes. However, it has limitations. Firstly, the use of explicit and implicit mixing, such as intrinsic decomposition and neural tone-mapping module $\mathcal{T}$, may reduce training and rendering speed. Secondly, in the nearly pure noise regions with extremely low signal-to-noise ratios (SNR)~\cite{johnson2006signal}, our rendering quality will degrade. For example, our result shown in Fig. ~\ref{fig:results1} (\textit{path}) fails to recover the details in shadowed regions. Enhancing rendering quality in such condition is a future optimization direction.

\vspace{-1ex}
\section{CONCLUSION}
We propose LL-Gaussian, a novel framework for low-light scene reconstruction and pseudo normal-light novel view synthesis from sRGB images. It introduces a robust low-light Gaussian initialization model, dual-branch Gaussian decomposition for intrinsic and transient modeling, and a physically guided unsupervised optimization. LL-Gaussian effectively handles noise, low dynamic range, and unstable initialization, enabling real-time, high-fidelity rendering without RAW inputs or exposure metadata. Extensive experiments show that it outperforms all baselines in quality, noise robustness, and training speed, making it practical for real-world deployment.

\bibliographystyle{ACM-Reference-Format}
\bibliography{sample-base}

\clearpage

\setcounter{page}{1}
\setcounter{section}{0}
\setcounter{figure}{0}
\setcounter{table}{0}
\renewcommand{\thesection}{\arabic{section}}

\section*{Supplementary Materials}
\vspace{0.5cm}

\section{Overview}
With in the supplementary, we provide:
\begin{itemize}
    \item LLGIM Algorithm Description in Sec. \ref{sec:No1}
    \item Affine alignment in luminance channel in Sec. \ref{sec:No2}
    \item Implementation details in Sec. \ref{sec:No3}
    \item Comparison with LLIE+Scaffold-GS methods in Sec. \ref{sec:No4}
    \item Comparison on LLNeRF Dataset in Sec. \ref{sec:No5}
    \item More qualitative results in Sec. \ref{sec:No6}
\end{itemize}

\section{LLGIM Algorithm Description}
\label{sec:No1}
The full version of the Low-Light Gaussian Initialization Module (LLGIM) is detailed in Algorithm~\ref{alg:llgim}.

\section{Affine Alignment in Luminance Channel}
\label{sec:No2}
To ensure a fair evaluation of unsupervised low-light scene enhancement methods, we per-process the enhancement rendering outputs by affine alignment in the luminance channel, which mitigates the influence of illumination discrepancies between enhanced results and pseudo normal-light GT images. First, we convert both the enhanced results and GT images from sRGB color space to LAB color space to decouple luminance information from chromatic components. Following RawNeRF~\cite{mildenhall2022nerf}, for each output and
the ground truth clean image, we process as the following procedure :
\begin{equation}
    a=\frac{\overline{xy}-\overline{x}\ \overline{y}}{\overline{x^2}-\overline{x}^2}=\frac{\text{Cov}(x,y)}{\text{Var}(x)},b=\overline{y}-a\overline{x}.
\end{equation}
where $x$ is the luminance channel of ground truth and the luminance channel to be matched is y, $\overline{x}$ is the mean of $x$.This process is the least-squares fit of an affine transform $ax+b\approx y$. During testing, we align the enhanced luminance channel $y$ using the affine transformation $(y - b) / a$, and then convert the aligned outputs back to the sRGB color space to calculate the evaluation metrics.

\begin{algorithm}[t]
\caption{\textsc{Low-Light Gaussian Initialization Module (LLGIM)}}
\label{alg:llgim}
\begin{algorithmic}[1]
\State \textbf{Input:} Low-light image set $\mathcal{I} = \{I_i\}$
\State \textbf{Output:} Pruned 3D Gaussian anchors $\mathcal{A}^*$
\vspace{0.3em}

\State \textit{\textbf{1. Dense Point Cloud Injection}}
\State $\mathcal{P} \gets \text{DUSt3R}(\mathcal{I})$ \Comment{Obtain dense point cloud}
\State Construct voxel grid $\mathcal{V}$ and generate anchor candidates $\mathcal{A}^{(0)}$
\vspace{0.3em}

\State \textit{\textbf{2. Distance-Adaptive Stochastic Pruning}}
\State Initialize pruning threshold $\tau^{(0)}$ and temperature $\beta$
\For{$t = 0$ to $T$}
    \For{each anchor $a_k \in \mathcal{A}^{(t)}$}
        \State Compute $d_{\text{min}}(a_k)$ \Comment{Min distance to other anchors}
        \State Compute probability:
        \[
            P(a_k) = \min\left(1, \frac{d_{\text{min}}(a_k)}{\tau^{(t)}} + \epsilon \right)
        \]
        \State Sample retention with $\text{Bernoulli}(P(a_k))$
    \EndFor
    \State $\mathcal{A}^{(t+1)} \gets$ retained anchors
    \State Update threshold:
    \[
        \tau^{(t+1)} = \tau^{(t)} \cdot \exp\left( \beta \cdot \frac{|\mathcal{A}^{(t)}|}{|\mathcal{A}^{(0)}|} \right)
    \]
\EndFor
\State $\mathcal{A}^* \gets \mathcal{A}^{(T)}$
\vspace{0.3em}

\State \textit{\textbf{3. Depth-Guided Warm-up Refinement }}
\State Obtain monocular depth prior $D^{\text{mono}}_k$ from a pre-trained estimator (e.g., Depth Anything V2)
\For{each view $k$}
    \State Render predicted depth $\hat{D}_k$ from current anchor set $\mathcal{A}^*$
    \State Compute PCC loss:
    \[
    \mathcal{L}_{\text{depth}} = 1 - \frac{\text{Cov}(\hat{D}_k, D_k^{\text{mono}})}{\sigma\{\hat{D}_k\} \cdot \sigma\{D_k^{\text{mono}}\}}
    \]
    \State Backpropagate $\mathcal{L}_{\text{depth}}$ to refine anchor positions
\EndFor
\State Update $\mathcal{A}^*$ via gradient-based*
\State \textbf{return} Final anchors $\mathcal{A}^*$
\end{algorithmic}
\end{algorithm}

\section{Implementation Details.} 
\label{sec:No3}
We build our method upon Scaffold-GS~\cite{lu2024scaffold}. We train our models for 8k iterations across all scenes and use the same loss function. We set $r,\ \tau^{(0)},\ \beta=1$ to stochastic prune the DUSt3R-initialized point cloud. For decomposition components' decoders, we use the Adam optimizer with an initial learning rate of 4.0e-1. The initial learning rates for offset for each intrinsic gaussians and transient gaussians are set to 1.0e-3 and 5.0e-3, respectively, other settings are the same as those of Scaffold-GS. Additionally, we perform experiments on one NVIDIA TESLA V100 GPU for fair comparisons.

\section{Comparison with LLIE+Scaffold-GS Methods.}
\label{sec:No4}
\begin{figure*}
\begin{center}
\includegraphics[width=0.95\linewidth]{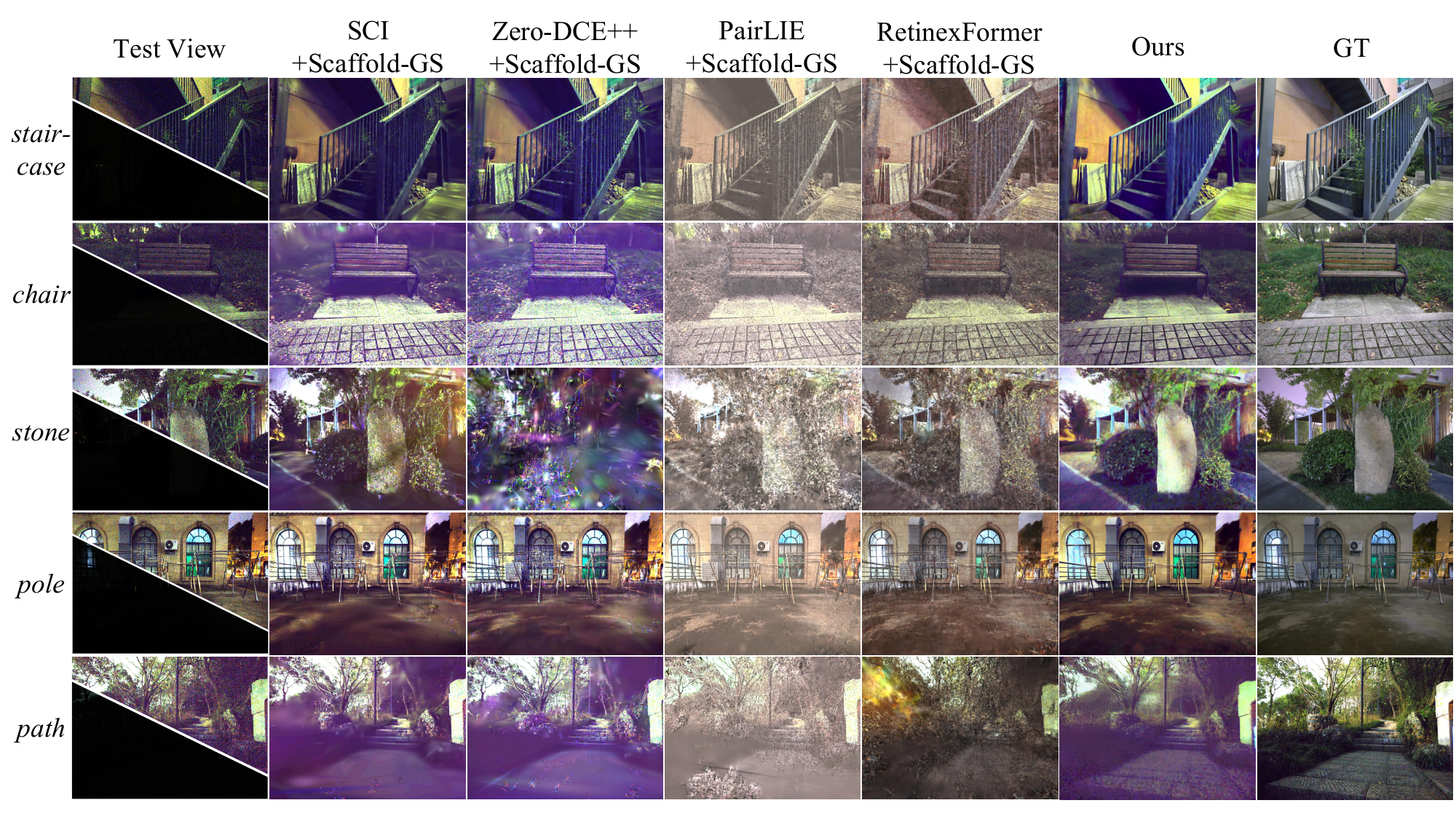}
\end{center}
\caption{Comparison Results with LIE+Scaffold-GS on our LLRS Dataset. Due to the partial compromise of 3D consistency in low-light scene data by LLIE methods, significant artifacts and noise Gaussians are generated during Scaffold-GS reconstruction of enhanced multi-view images. We significantly outperform all LLIE + Scaffold-GS methods.}
\label{fig:supp_results_LLIE}
\end{figure*}

It is a natural approach to first enhance the input multi-view dataset using  low-light image enhancement (LLIE) methods and then reconstruct normal-light scenes through 3D reconstruction algorithms in a multi-stage pipeline. To comprehensively validate the effectiveness of our proposed end-to-end LL-Gaussian algorithm, we compare our method with the LLIE+Scaffold-GS approach. Specifically, we first apply SOTA LLIE methods including SCI~\cite{ma2022toward}, Zero-DCE++~\cite{Zero-DCE++}, PairLIE~\cite{fu2023learning}, and RetinexFormer~\cite{retinexformer} to enhance the low-light multi-view images. The enhanced multi-view images are then processed using COLMAP~\cite{schonberger2016structure} to obtain initial point clouds and camera poses. Finally, we use the novel view synthesis is performed using the original Scaffold-GS. As shown in Fig. ~\ref{fig:supp_results_LLIE} and Table ~\ref{tab:supp_results_LLIE}, our LL-Gaussian achieves significantly superior results compared to the LLIE + Scaffold-GS approach. Although existing LLIE methods leverage data priors through extensive training on low-light/normal-light datasets, they face inherent limitations: 1) Training data cannot cover all real-world low-light scenarios; 2) Variations in camera equipment introduce different noise patterns. These factors make it challenging for LLIE methods to preserve 3D consistency in scene reconstruction and completely eliminate noise interference, particularly under extreme low-light conditions (with lower signal-to-noise ratio). Consequently, the LLIE pre-processing introduces substantial interference signals to the Scaffold-GS reconstruction, resulting in significant artifacts and noise-corrupted Gaussians as demonstrated in Fig. ~\ref{fig:supp_results_LLIE}. In contrast, our LL-Gaussian, by adopting an end-to-end approach that achieves joint optimization of reconstruction and enhancement, can effectively restore valid information from low-light data while preventing the disruption of the original 3D consistency.

\begin{table}[tb]
\centering
\caption{Quantitative comparisons with LLIE+Scaffold-GS methods on our LLRS dataset.  The \colorbox{red!30}{first},  \colorbox{orange!30}{second} and \colorbox{yellow!30}{third} best-performing methods highlighted. Since the LLIE methods We significantly outperform all LLIE + Scaffold-GS methods.}
\renewcommand\arraystretch{0.9}
\setlength{\tabcolsep}{1.2pt}
{\footnotesize
\begin{tabular}{c c c ccc }
\toprule
\multirow{2}{*}{\textbf{Methods}}  & \multicolumn{3}{c}{\textbf{LLRS Dataset}} \\
\cline{2-4}
\rule{0pt}{10pt}
 &  SSIM$\uparrow$ & LPIPS$\downarrow$ & PSNR$\uparrow$ \\
\midrule
PairLIE + Scaffold-GS      & \second{0.374} & \best{0.576} & \second{17.39} \\
RetinexFormer + Scaffold-GS & 0.342 & \second{0.587} & \third{17.38} \\
SCI + Scaffold-GS           & \third{0.359} & 0.652 & 17.18 \\
Zero-DCE++ + Scaffold-GS    & 0.290 & 0.667 & 15.77 \\
\midrule
ours          & \best{0.414} & \third{0.595} & \best{18.29} \\
\bottomrule
\end{tabular}
}
\label{tab:supp_results_LLIE}
\end{table}

\section{Comparison on LLNeRF Dataset.} 
\label{sec:No5}
\begin{figure*}
\begin{center}
\includegraphics[width=\linewidth]{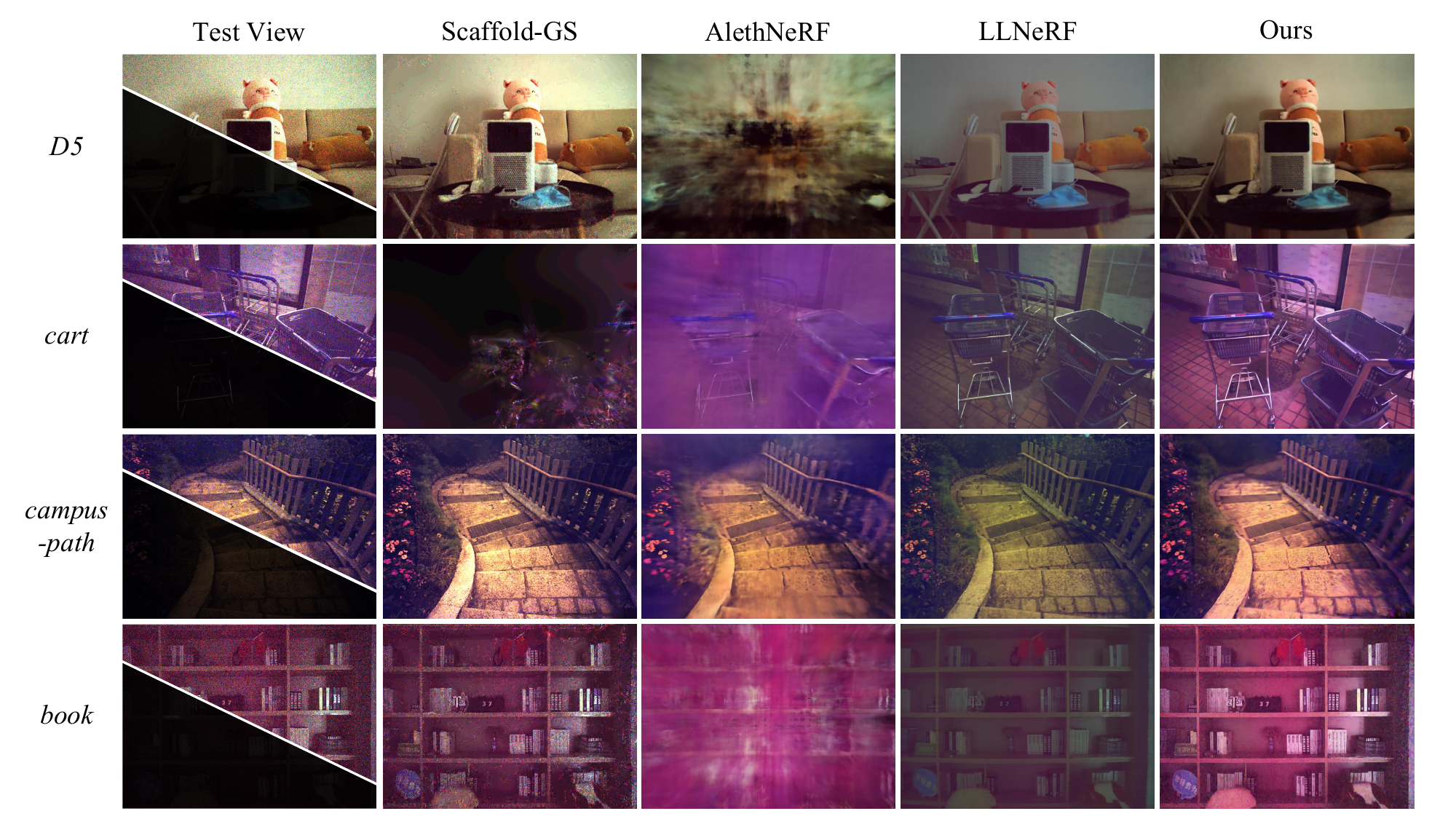}
\end{center}
\caption{Comparison Results on LLNeRF Dataset( "D5", "cart", "campus-path", "book" ).}
\label{fig:supp_results3}
\end{figure*}

The LLNeRF dataset collects challenging low-light noisy scenes, we choose four challenge scenes ( "D5", "cart", "campus-path", "book" ) with different types for comparison. Due to the absence of normal-light reference images precludes comprehensive quantitative evaluation, our comparison focuses on qualitative comparisons, with visual results presented in Figure~\ref{fig:supp_results3}. Compared to the baseline methods, our method exhibits stronger noise robustness and superior detail restoration capability.

\section{More Qualitative Results}
\label{sec:No6}
\begin{figure*}
\begin{center}
\includegraphics[width=0.95\linewidth]{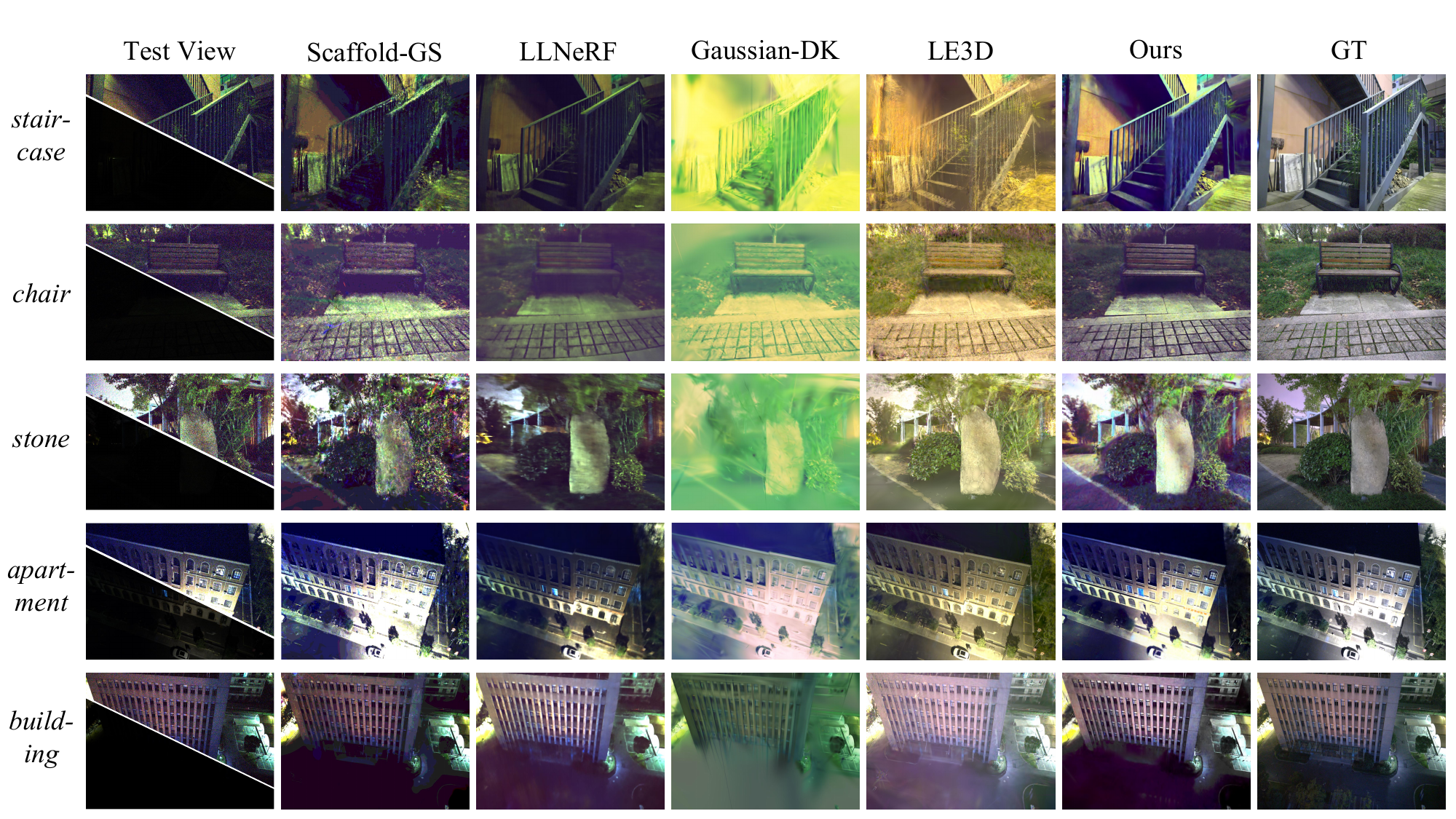}
\end{center}
\caption{Comparison Results on our LLRS Dataset ( "staircase", "chair", "stone", "apartment", "building" ).}
\label{fig:supp_results1}
\end{figure*}

\begin{figure*}
\begin{center}
\includegraphics[width=\linewidth]{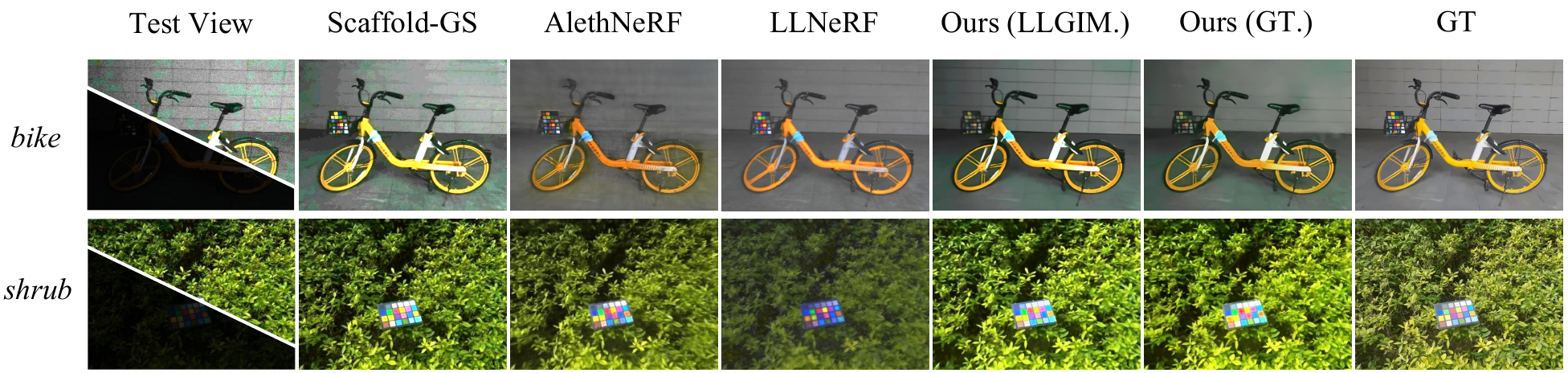}
\end{center}
\caption{Comparison Results on AlethNeRF Dataset( "bike", "shrub" ).}
\label{fig:supp_results2}
\end{figure*}

Fig. ~\ref{fig:supp_results1} show the qualitative results on the left five scenes( "staircase", "chair", "stone", "apartment", "building" ) of our LLRS dataset. Fig. ~\ref{fig:supp_results2}  show the qualitative results on the left two scenes( "shrub", "bike" ) of LOM dataset.

\end{document}